\pdfoutput=1

\documentclass[runningheads]{llncs}
\usepackage{graphicx, caption}
\usepackage[bookmarks=false]{hyperref}
\usepackage[caption=false]{subfig}
\usepackage{cite}
\usepackage{booktabs}

\begin{document}

\title{TSXplain: Demystification of DNN Decisions for Time-Series using Natural Language and Statistical Features}
%
%
\author{Mohsin Munir\inst{1,2} \and
Shoaib Ahmed Siddiqui\inst{1,2} \and
Ferdinand K{\"u}sters\inst{3} \and
Dominique Mercier\inst{1,2} \and
Andreas Dengel\inst{1,2} \and
Sheraz Ahmed\inst{2}}
\authorrunning{M. Munir et al.}
%
\institute{Technische Universit{\"a}t Kaiserslautern, Kaiserslautern 67663, Germany \and
German Research Center for Artificial Intelligence (DFKI) GmbH, Kaiserslautern 67663, Germany \\
\email{firstname.lastname@dfki.de}\\
\and
IAV GmbH, Gifhorn 38518, Germany\\
\email{ferdinand.kuesters@iav.de}}
\maketitle              
\begin{abstract}
Neural networks (NN) are considered as black-boxes due to the lack of explainability and transparency of their decisions. This significantly hampers their deployment in environments where explainability is essential along with the accuracy of the system. Recently, significant efforts have been made for the interpretability of these deep networks with the aim to open up the black-box. However, most of these approaches are specifically developed for visual modalities. In addition, the interpretations provided by these systems require expert knowledge and understanding for intelligibility. This indicates a vital gap between the explainability provided by the systems and the novice user. To bridge this gap, we present a novel framework i.e. Time-Series eXplanation (TSXplain) system which produces a natural language based explanation of the decision taken by a NN. It uses the extracted statistical features to describe the decision of a NN, merging the deep learning world with that of statistics. The two-level explanation provides ample description of the decision made by the network to aid an expert as well as a novice user alike. Our survey and reliability assessment test confirm that the generated explanations are meaningful and correct.  
We believe that generating natural language based descriptions of the network's decisions is a big step towards opening up the black-box.

\vspace{-5mm}
\end{abstract}
\vspace{-3mm}
\section{Introduction}

Deep neural networks (DNN) have become ubiquitous these days. They have been successfully applied in a wide range of sectors including automotive~\cite{kang2016intrusion}, government~\cite{trippi1992neural}, wearable~\cite{ordonez2016deep}, dairy~\cite{tsForDairy}, home appliances~\cite{singh2018deep}, security and surveillance~\cite{kushwaha2018deep}, health~\cite{tsForHealthcare} and many more, mainly for regression, classification, and anomaly detection problems~\cite{cai2018high, munir2018data, AnomalyDetectionPhD2014, deepAnT}. 
The neural network's capability of automatically discovering features to solve any task at hand makes them particularly easy to adapt to new problems and scenarios. 
However, this capability to automatically extract features comes at the cost of lack of transparency/intelligibility of their decisions.
Since there is no clear indication regarding why the network reached a particular prediction, these deep models are generally referred to as a black-box~\cite{tsviz}. 

It has been well established in the prior literature that explanation of the decision made by a DNN is essential to fully exploit the potential of these networks~\cite{saad2007neural, andrews1995survey}. 
With the rise in demand for these deep models, there is an increasing need to have the ability to explain their decisions. 
For instance, big industrial machines cannot be powered down just because a DNN predicted a high anomaly score. It is important to understand the reason for reaching a particular decision, i.e. why the DNN computed such an anomaly score. 
Significant efforts have been made in the past in order to better understand these deep models~\cite{tsviz,bach2015pixel,yosinskiunderstanding,simonyan2013deep}. 
Since most of the prior literature is validated specifically for visual modalities, therefore, a very common strategy is to visualize the learned features~\cite{yosinskiunderstanding, simonyan2013deep, kuo2018interpretable, zeiler2014visualizing}. Visual representation of filters and learned features, which might provide important hints to the expert, is already a good step in this direction, but might not be intelligible to the end user. Adequate reasoning of the decision taken increases the user's confidence in the system. 

The challenges pertaining to interpretability and visualization of deep models developed for time-series analysis are even more profound. 
Directly applying the interpretation techniques developed specifically for visual modalities are mostly uninterpretable in time-series data, requiring human expertise to extract useful information out of these visualizations~\cite{tsviz}. Therefore, we aim to bridge this gap by generating explanations based on the concrete knowledge extracted out of the statistical features for the decision made by the network, assisting both novice and expert users alike. Statistical features are considered to be transparent which ameliorates the process of gaining user trust and confidence. 

\begin{figure*}[t] 
\centering
\includegraphics[width=0.85\textwidth]{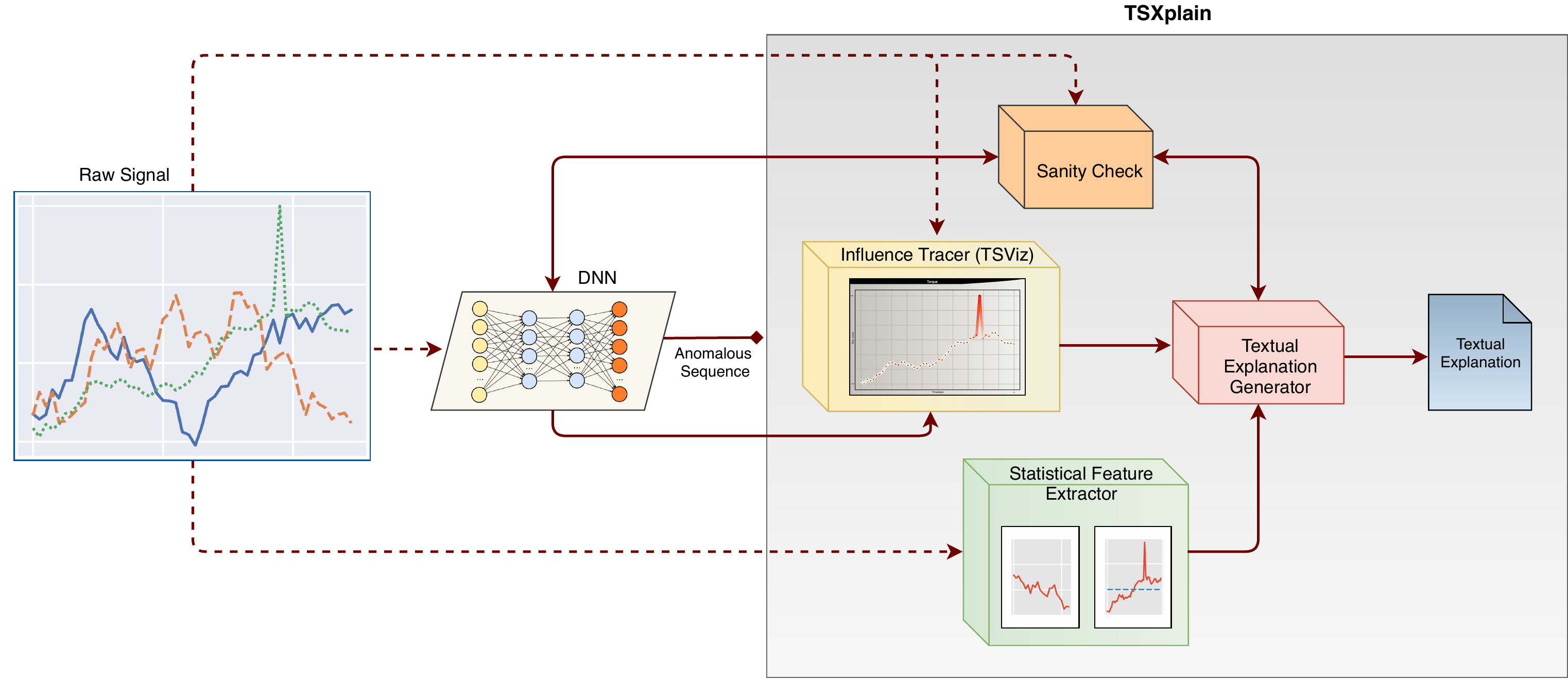}
\caption{All the modules and their connections are shown in this TSXplain system diagram. The system is only activated for anomalous time-series.}
\label{fig:sysOverview}
\vspace{-4mm}
\end{figure*}

This paper presents a novel approach for generating natural language explanations of the classification decisions made by a DNN (whole pipeline is shown in Fig.~\ref{fig:sysOverview}). We take a pre-trained DNN and find the most salient regions of the input that were mainly leveraged for a particular prediction. Important data points which contributed towards the final network decision are tracked using the influence computation framework.
These points are then combined with different statistical features, which are further used to generate natural language explanations.
We specify a confidence to the generated explanation by accessing its reliability through sanity check.
Our experiments on the classification datasets and a survey show that the generated natural language explanations help in better understanding the classification decision. 
To the best of the authors' knowledge, it is the very first attempt to generate natural language explanations for classification task on time-series data.


\vspace{-2mm}
\section{Related Work} \label{sec:related}

Over the past few years, there have been numerous advancements in the field of network interpretation and visualizations. For understanding a DNN on an image classification task, Zeiler and Fergus (2014)~\cite{zeiler2014visualizing} proposed a technique for the visualization of deep convolutional neural networks. Their visualization reveals features in a fully trained network. Highlighting the input stimuli which excites individual feature maps at different layers, helps in understanding what the network has learned. 
Yosinski et al. (2015)~\cite{yosinskiunderstanding} introduced DeepVizToolbox which helped in understanding how neural network models work and which computations they perform at the intermediate layers of the network at two different levels. 
This toolbox visualizes the top images for each unit, forward activation values, deconvolutional highlighting, and preferred stimuli. Simonyan et al. (2013)~\cite{simonyan2013deep} also presented an approach to visualize the convolutional neural networks developed for image classification. 
Their visualization provides image-specific class saliency maps for the top predicted class.

Bau et al. (2017)~\cite{bau2017network} introduced a framework to interpret the deep visual representations and quantify the interpretability of the learned CNN model. First, they gather a broad set of human labeled visual concepts and then gather the response of hidden variables to known concepts.
To understand a neural network, Mahendran and Vedaldi (2015)~\cite{mahendran2015understanding} highlight the encoding learned by the network through inversion of the image representations. They also study the locality of the information stored in the representations. Melis et al. (2018)~\cite{melis2018towards} designed self-explaining models where the explanations are intrinsic to the model for robust interpretability of a network.
Bach et al. (2015)~\cite{bach2015pixel} introduced an approach to achieve pixel-level decomposition of an image classification decision. They generate a heatmap for a well-classified image which segments the pixels belonging to the predicted class. 

Current DNN visualizations and interpretations only help an expert to understand and improve the overall process. However, they still lack the actual reasoning why a particular decision has been taken by the learned model. There have been some work in the domain of image captioning where visual attributes are leveraged to support the DNN decision. Guo et al. (2018)~\cite{guo2018neural} proposed a textual summarization technique of image classification models. They train a model with the image attributes which are used to support the classification decision.
In the same domain, Hendricks et al. (2016)~\cite{hendricks2016generating} proposed a model which predicts a class label and explains the reason of the classification based on the discriminating properties of the visual objects. Kim et al. (2018)~\cite{kim2018textual} introduced a textual explanation system for self-driving vehicles. They generate introspective explanations to represent the causal relationships between the system's input and its behavior which is also the target of our study. 
In most of the aforementioned techniques, a neural network black-box is further used to generate descriptions and explanations of another neural network. Despite being a promising direction, this introduces another level of opaqueness into the system.  

In the domain of relation extraction, Hancock at al. (2018)~\cite{hancock2018training} proposed a supervised rule-based method to train classifiers with natural language explanations. In their framework, an annotator provides a natural language explanation for each labeling decision. 
Similar work has been presented by Srivastava et al. (2017)~\cite{srivastava2017joint}, but they jointly train a task-specific semantic parser and classifier instead of a rule-based parser. These systems, however, rely on a labelled set of training examples that are not available in most of the real-world applications.

\vspace{-2mm}
\section{Methodology} \label{sec:method}

Different visualization and interpretation techniques developed specifically to understand deep models aid an expert in understanding the learning and decision-making process of the network. However, the provided interpretation / visualization cannot be readily understood by a novice user. It is up to the user to draw conclusions about the network's decision with the help of the available information. 
Many of the existing techniques use a separate deep network that is trained for the generation of explanations using the primary model~\cite{guo2018neural,hendricks2016generating,kim2018textual}. These explanations still suffer from lack of transparency, as they are also generated by a deep model. 
Therefore, we approach the problem of generating explanations in a way which significantly improves the intelligibility of the overall process. 
We leverage the statistical time-series features to provide a concrete natural language based explanation of an anomalous sequence. These features also help in gaining user's trust because of their lucid nature.
The proposed system is composed of different modules as highlighted in Fig.~\ref{fig:sysOverview}. The raw input is first passed on to DNN for classification. If the sequence is classified an anomalous, the whole TSXplain system is activated which is composed of four modules, namely \textit{influence tracer (TSViz)}~\cite{tsviz}, \textit{statistical feature extractor, sanity check,} and \textit{textual explanation generator}. 
The \textit{influence tracer} is employed to discover the most salient regions of the input (Section~\ref{sec:tsviz}). The \textit{statistical feature extractor} module extracts different statistical features from the sequence (Section~\ref{sec:featureExtractor}). The results from previous two modules are passed onto the \textit{textual explanation generator} module in order to come up with a natural language description of the encountered anomaly (Section~\ref{sec:description}). Furthermore, we introduce a \textit{sanity check} module to get a coarse estimate of the system's confidence regarding the generated explanation (Section~\ref{sec:sanityCheck}). 

\vspace{-2mm}
\subsection{Influence Tracer (TSViz)} \label{sec:tsviz}

Influence tracer module is based on the TSViz framework\cite{tsviz}. The proposed influence tracing algorithm can be used to trace the influence at several different levels. However, we only consider the main influence for our method i.e. the influence of the input on the output. These influences can be effectively computed as the gradients of the output $y$ w.r.t. the input components $x_j$:
\begin{equation}
I_{\mathrm{input}}^{j} = \left|\frac{\partial y}{\partial x_j}\right|
\end{equation}
The absolute value of the gradient is used here, as only the magnitude of the influences is of relevance to us. Once the influences have been computed, we use the max-min scaling to scale the values for visualization and further processing as specified by:
\begin{equation} 
	I_{\mathrm{input, scaled}}^{j} = \frac{I_{\mathrm{input}}^{j} - \min\limits_{j} I_{\mathrm{input}}^{j}} {\max\limits_{j} I_{\mathrm{input}}^{j} - \min\limits_{j} I_{\mathrm{input}}^{j}}
\end{equation}

The computed influence values are visualized on top of the original signals to provide a hint regarding the encapsulated information. 
Siddiqui et al. (2018)~\cite{tsviz} discussed the problem of extremely confident predictions for the influence tracing algorithm and suggested a remedy to overcome this issue by imposing regularization on top of the activations itself when training the network. The new objective can be written as:


\noindent where $\Phi$ defines the mapping from the input to the output space, $\mathcal{W}$ encapsulates all the parameters of the network, $z^L$ denotes the activation values of the last layer before application of the sigmoid layer, and $\lambda$ and $\beta$ denotes hyperparameters controlling the contribution of the regularization terms and the empirical risk. We use the same modified objective to train our network in order to avoid extremely confident predictions.

The \textit{influence tracer} module consumes both the DNN model as well as the raw input. Then, it performs the backward pass through the network from output to input in order to obtain these influence values. The output from this module is consumed by the \textit{textual explanation generator} module (Section~\ref{sec:description}). 

\vspace{-1.3mm}
\subsection{Statistical Feature Extractor} \label{sec:featureExtractor}

This module extracts different statistical features from the input sequence. 
Since we are dealing with sequential data comprising of time-series, we calculate different point-wise as well as sequence-wise features. These features include, but not limited to, lumpiness, level shift, KL score, number of peaks, and ratio beyond r-sigma (explained below). These features have been previously used by Bandara et al. (2017)~\cite{bandara2017forecasting} and have been proposed in~\cite{wang2006characteristic, hyndman2015large}. 

\begin{enumerate}
    \item\textit{Lumpiness:} Initially, daily seasonality from the sequence is removed by dividing it into blocks of \textit{n} observations. Variance of the variances across all blocks is computed which represent the lumpiness of the sequence.
    \item\textit{Level shift:} The sequence is divided into \textit{n} observations and the maximum difference in mean between consecutive blocks is considered as the level shift. It highlights the block which is different from the rest of the sequence.
    \item\textit{KL score:} To calculate this score, the sequence is divided into consecutive blocks of \textit{n} observations. This score represents the maximum difference in Kullback-Leibler divergence among consecutive blocks. A high score represents high divergence. 
    \item\textit{Number of peaks:} This feature identifies the number of peaks in a sequence. The sequence is smoothed by a Ricker wavelet for widths ranging from $1$ to $n$. It detect peaks with sufficiently high signal-to-noise ratio.
    \item\textit{Ratio beyond r-sigma:} It gives the ratio of data points which are \textit{r} standard deviations away from the mean of a sequence.
    \item\textit{Standard deviation:} This feature represents the standard deviation of a sequence.
\end{enumerate}

\begin{figure}[b!] 
\centering
{\includegraphics[width=0.99\textwidth]{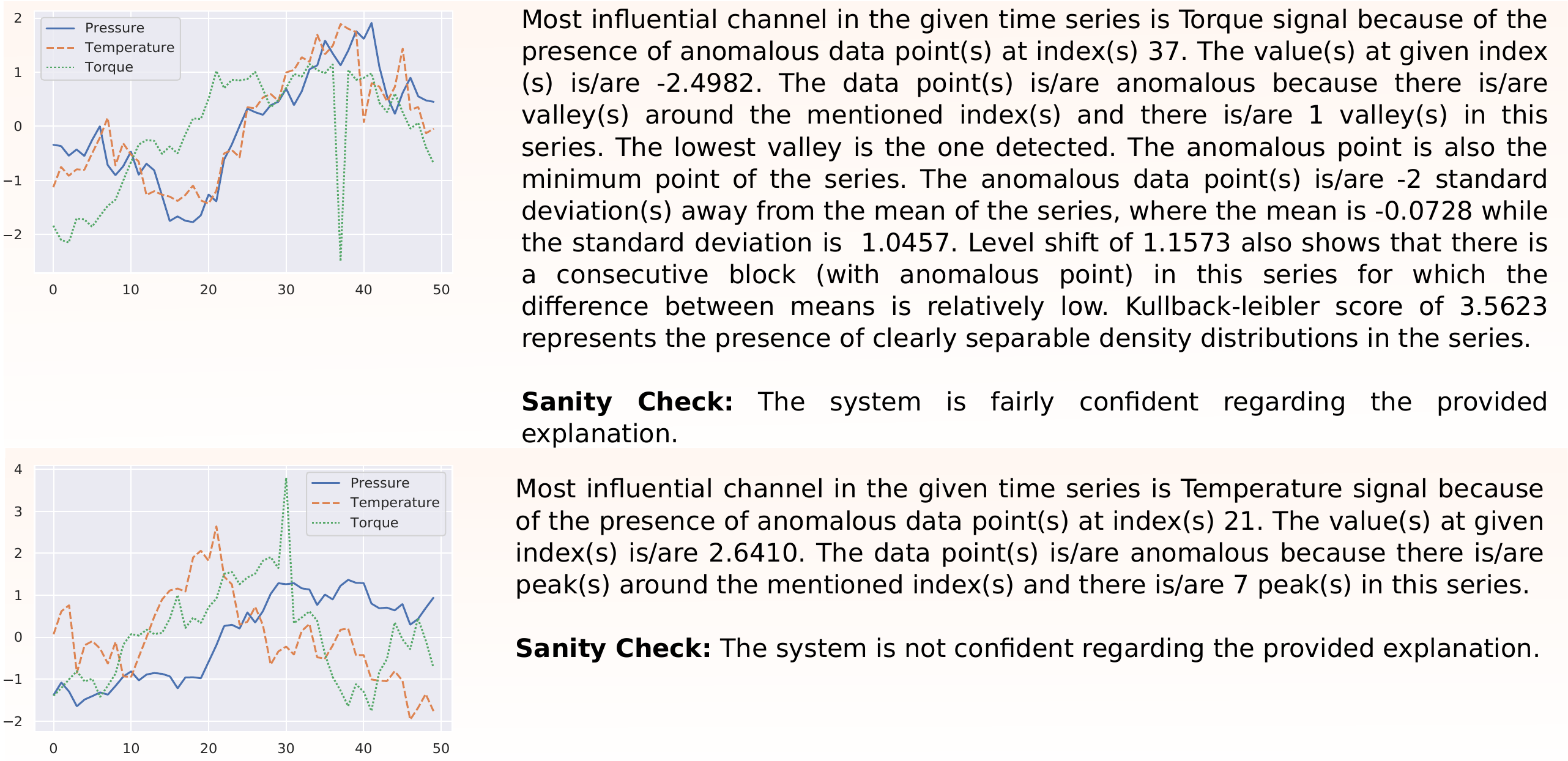}}
\vspace{-2mm}
\caption{Channel-wise natural language explanations generated for expert user. Second example shows a failure case.
}
\label{text}
\end{figure}

In addition to the above mentioned sequence-wise features, we also use point-wise features including peak, valley, maximum, minimum, highest spike, and lowest valley etc. 
The fusion of sequence-wise and point-wise features provides vivid characteristics of the highly influential data points. 

\vspace{-2mm}
\subsection{Textual Explanation Generator} \label{sec:description}

The influential points determined by the \textit{influence tracer} module along with the features computed by the \textit{statistical feature extractor} module are passed onto this module for the generation of the textual explanations of a given anomalous sequence. This module also receives input from the \textit{sanity check} module which allows the system to specify its confidence over the generated explanation. 

We designed a set of rules to incorporate a range of features. These rules are defined based on the statistical time-series features in a way which explains different characteristics of a given sequence. Based on the classification decision given by the network along with the time-series features, this module provides explanations of the data points which influence the network decision.
The explanations are generated channel-wise, so that anomalous data points in each channel can be highlighted. When multiple channels are available, the salient salient data points from each individual channel are passed onto this module.

\begin{figure}[b] 
\centering
\hspace{-1mm}
\subfloat[\small Anomalous sequence.]{\includegraphics[width=0.45\textwidth]{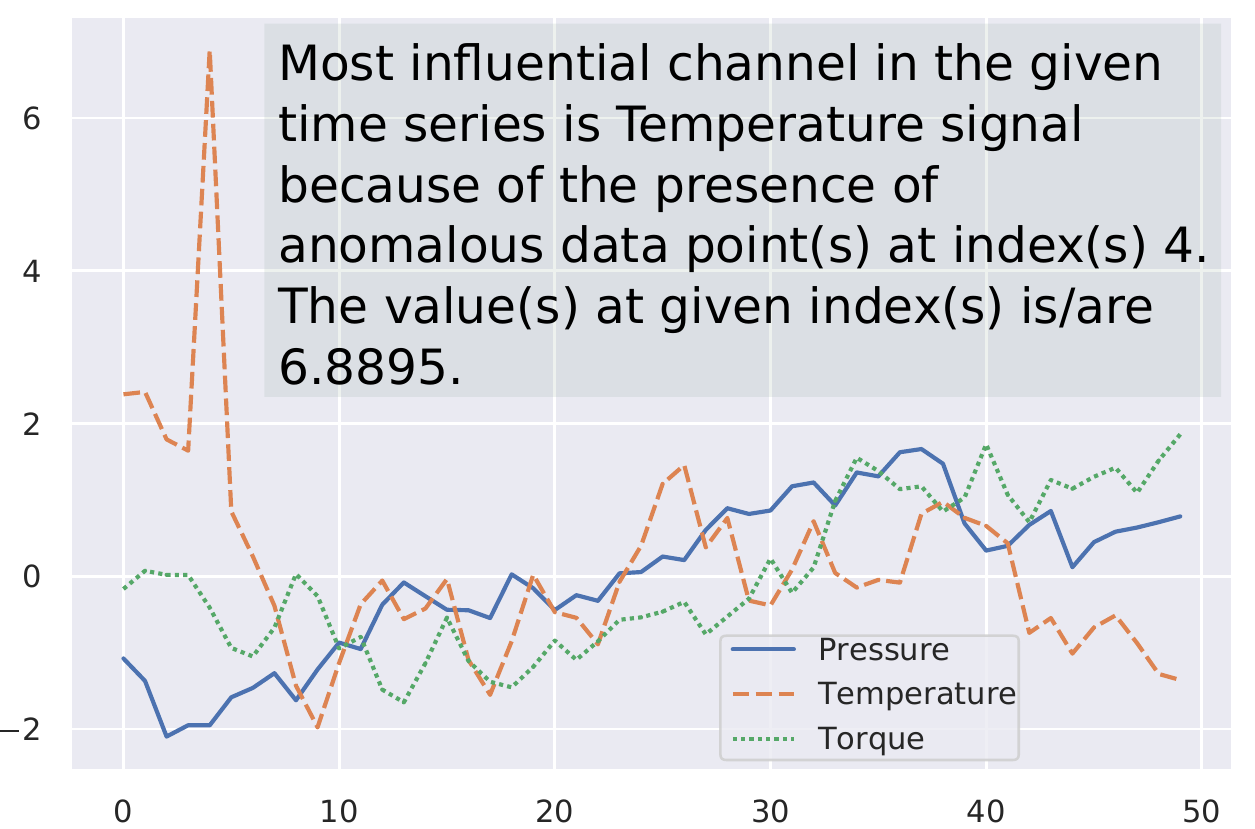}\label{fig:original}}
\hfil
\hspace{-1mm}
\subfloat[Effect of sanity check.]{\includegraphics[width=0.45\textwidth]{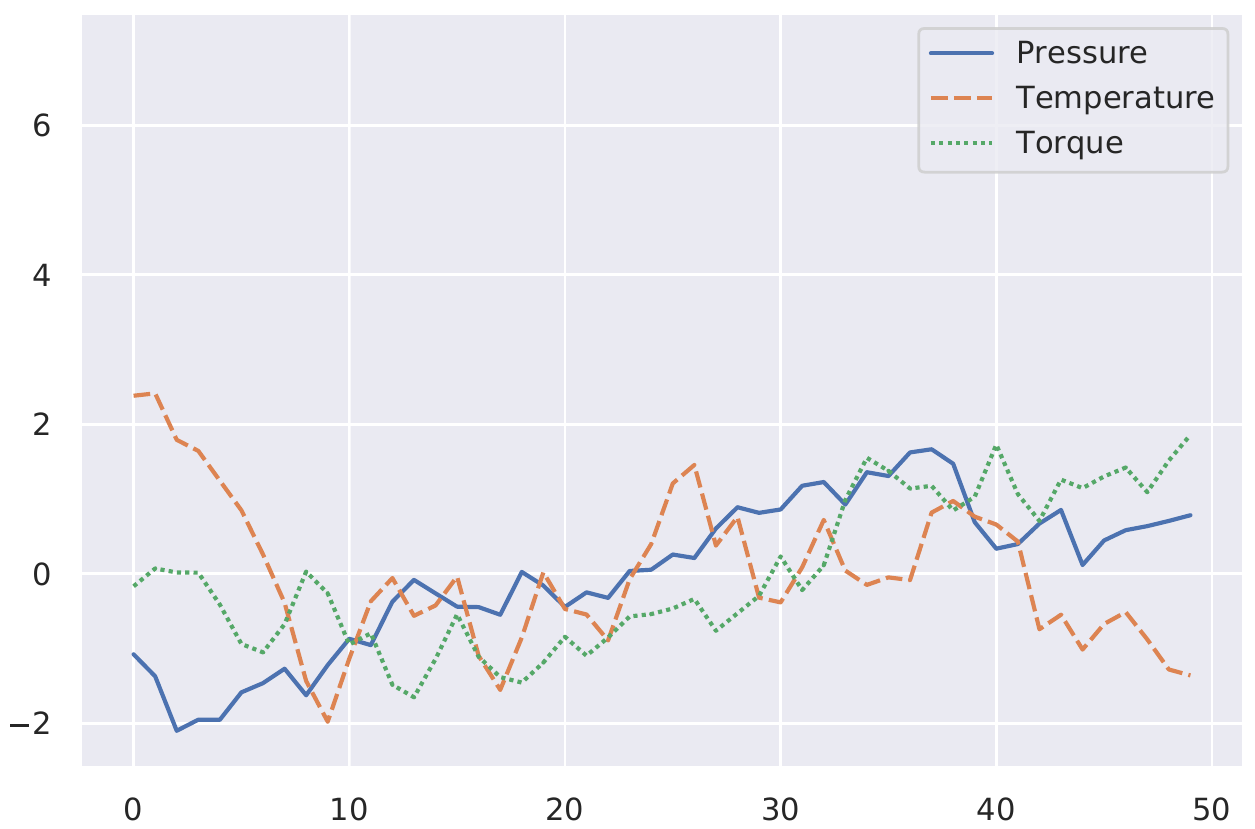}\label{fig:sanity}}
\vspace{-2mm}
\caption{ 
As a results of sanity check, the anomalous peak is suppressed and sequence in (a) is classified as normal in (b).
}
\label{fig:sanity_comparison}
\vspace{-5mm}
\end{figure}

We have defined two levels of abstraction for the textual explanations. 
First level of explanation is defined for the users who are not interested in detailed explanations or don't have enough knowledge of time-series data (novice user), 
but would like to get information regarding the most salient regions and channels of the input. 
The second level of explanation, on the other hand, is defined for the expert users. Such users are generally interested in knowing the details such as, why a network took a particular decision? 
Sample explanations for expert users are shown in Fig.~\ref{text}. The explanations shown in this figure clearly point out the anomalous channel in a given time-series sequence. Moreover, characteristics of anomalous data points and anomalous channel are also explained in the form of feature values. Fig.~\ref{fig:original} shows an example of simple explanation which is generated for a novice user.

\vspace{-1.5mm}
\subsection{Sanity Check} \label{sec:sanityCheck}
In order to assess the reliability of a given explanation, a simple sanity check is performed. The output of the \textit{textual explanation generator} module along with the original input is passed onto this module. The data point corresponding to the explanation are suppressed and this masked sequence is fed again to the network for inference. The masking is performed by linear interpolation between the last and the first retained points. 
If the removed points were indeed causal for the network prediction, we expect the prediction to flip. We use this sanity check to compute a confidence over the provided explanation. If we observe a flip in the prediction, we assign a high confidence to the provided explanation. On the other hand, if the prediction is retained, we assign a low confidence to the provided explanation. 
This check confirms that the generated explanations are referring to the data points which are actually contributing towards the classification decision taken by the network. Finally, the sanity check output is passed back to the \textit{textual explanation generator} module with the confidence information which is mentioned in Fig.~\ref{text} after the generated explanations. In this figure, the second example didn't observe any flip in the prediction after suppression of the deemed causal point (a failure case), resulting in low system confidence for the provided explanation. Fig.~\ref{fig:sanity_comparison} visualizes the process of sanity check on an anomalous sequence. In Fig.~\ref{fig:original}, the textual explanation highlights an anomalous data point in \textit{temperature} signal. The network prediction is flipped after suppressing the mentioned anomalous data point in \textit{temperature} signal as shown in Fig.~\ref{fig:sanity}.

\vspace{-2mm}
\section{Experimental Setup and Dataset} \label{sec:dataset}
\vspace{-1.5mm}
To classify a time-series as normal or anomalous, we trained a CNN model with three convolutional layers comprising of $16$, $32$ and $64$ filters respectively, with Leaky ReLU as the activation function, followed by a single dense layer. Since the focus of this paper is on generation of textual explanations, we selected the hyperparameters (e.g. number of layers, number of filters) based on our experience and did not invest any significant effort into hyperparameter optimization or model selection. To generate natural language explanations on time-series data, we used the following datasets:
\newline
\noindent\textbf{Machine Anomaly Detection~\cite{tsviz}:} It is a synthetic time-series classification dataset curated by Siddiqui et al. (2018)~\cite{tsviz}. This dataset comprises of $60000$ time-series with $50$ time-stamps each. Each sequence consists of three channels which represent values from pressure, torque, and temperature sensors.
The dataset contains point anomalies in the torque and temperature signals, while the pressure signal is kept intact. A sequence is labeled as anomalous if it contains a point anomaly. The dataset is split into $45000$ training sequences with $7505$ anomalous sequences, $5000$ validation sequences with $853$ anomalous sequences, and $10000$ test sequences with $1696$ anomalous sequences.

\noindent\textbf{Mammography~\cite{OpenML2013}:} This breast cancer screening dataset contains $11183$ time-series and it is commonly used for classification purposes. Anomalies at certain points/features make the whole time-series anomalous. The dataset is split into $8000$ training time-series with $186$ anomalous time-series, $1000$ validation time-series with $25$ anomalous time-series, and $2183$ test time-series with $49$ anomalous time series.

\noindent \textbf{NASA Shuttle~\cite{OpenML2013}:} This dataset describes radiator positions in a NASA shuttle with $9$ attributes. There are $46464$ time-series in this dataset. We split the dataset into $25000$ training time-series with $483$ anomalous time-series, $5000$ validation time-series with $102$ anomalous time-series, and $16464$ test time-series with $293$ anomalous time series.

\begin{figure}[t] 
\centering
\subfloat{\includegraphics[width=0.5\textwidth]{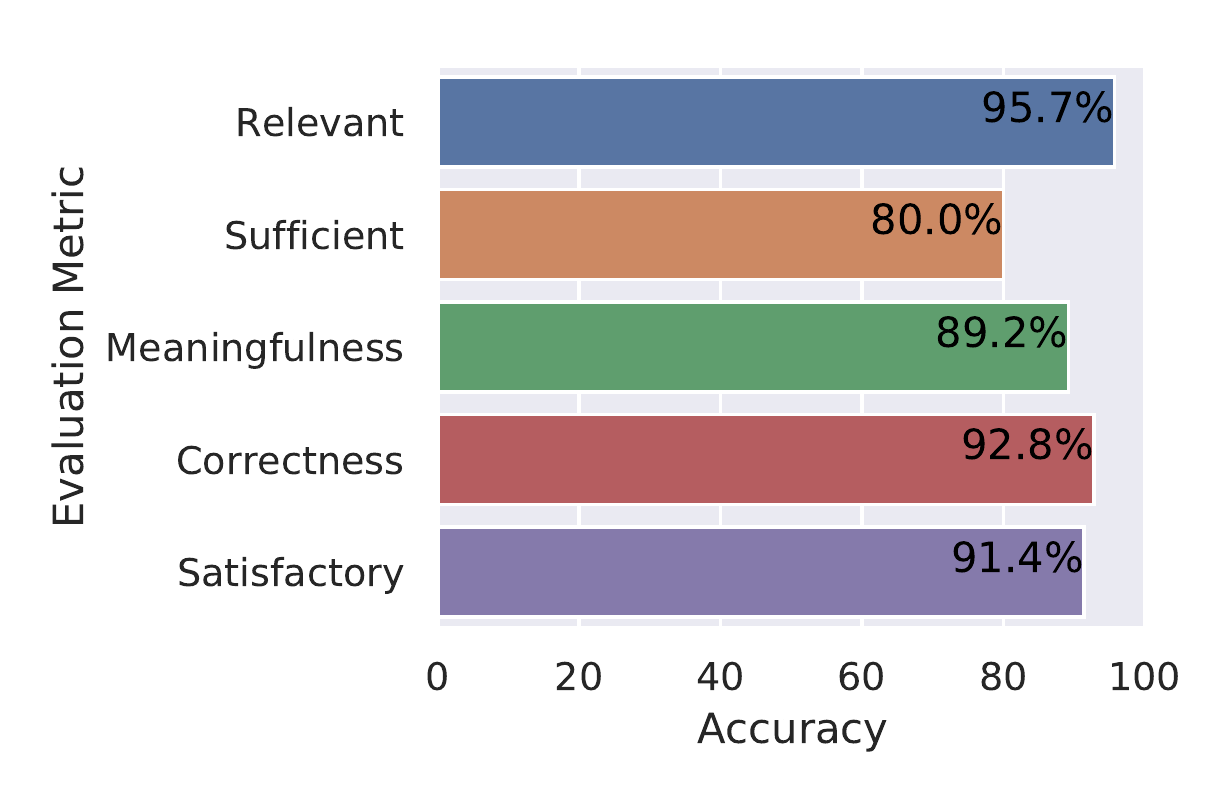}\label{fig:survey_expert}}
\subfloat{\includegraphics[width=0.5\textwidth]{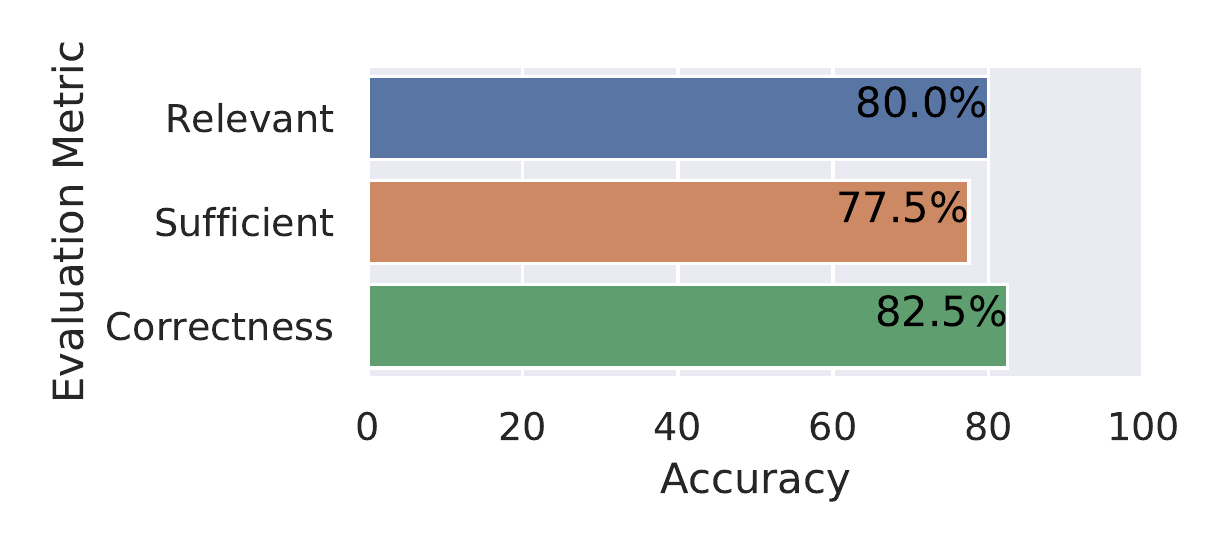}\label{fig:survey_novice}}
\vspace{-4mm}
\caption{Summary of evaluation done by expert (left) and novice (right) users on \textit{Machine Anomaly Detection} dataset explanations.}
\label{fig:survey}
\end{figure}

\section{Evaluation and Discussion} \label{sec:discussion}
In order to completely assess the relevance and correctness of a given explanation, we conducted a survey in which novice and expert users were asked to evaluate the generated explanations. We provided $20$ time-series from the \textit{Machine Anomaly Detection} dataset along with the generated explanations to the participants and asked questions related to whether the generation explanation was a) relevant, b) sufficient, c) meaningful, d) correct, and e) satisfactory from experts. Whereas, novice users were only questioned about (a), (b), and (d). $7$ expert and $6$ novice participants provided their binary (agree/disagree) feedback to the aforementioned questions.
By analyzing the accumulated feedback of the participants shown in Fig. \ref{fig:survey}, it is clear that most of the participants considered the provided explanations relevant, meaningful, and correct. Majority of the experts were satisfied with the reasoning of the NN decision provided in the explanation. Although, $20\%$ experts and $22.5\%$ novice participants thought that the provided explanation is not sufficient.



In this study, we are trying to infer the causality through the provided explanations, so it is also important to assess the reliability of the generated explanations. Therefore, we introduced the sanity check module in the system pipeline to obtain a measure of confidence over the provided explanations (as explained in Section~\ref{sec:sanityCheck}). We also computed this confidence estimate over the entire test set of \textit{Machine Anomaly Detection} dataset in order to get an impression regarding the overall reliability of the generated explanations. The cumulative results are presented in Table~\ref{table:sanity}. Since it is important to compute these statistics only over examples where the classification from the network was correct, we were left with $1511$ out of $1696$ total anomalous sequences in the test set. In the first setup, a masked sequence is generated by suppressing the exact data points for which the explanations have been generated by the \textit{textual explanation generator} module. Since we are suppressing the exact point, we represent this setup with a window size of one. In the second setup, a sequence is masked with a window size of three covering one preceding and one following value in order to cover up any minor misalignment of the most salient region highlighted by the \textit{textual explanation generator} module. In this case, a total of three data points were suppressed. We represent this setup with a window size of three. The results shown in Table~\ref{table:sanity} indicate that for $73.0\%$ of the anomalous sequences, the predictions were flipped by masking out the exact data points highlighted by the explanation module. When we relaxed the sanity check criteria to a window size of three, the percentage of flipped sequences rose up to $87.3\%$. This high success rate makes it evident that in most of the cases, plausible explanations for the predictions made by the network could be provided. However, it is important to note that this experiment does not strongly imply causality.

\tabcolsep=0.15cm
\begin{table}[t]
\tiny
\centering
\caption{Effect of masking the data points in \textit{Machine Anomaly Detection} dataset which are relevant for the explanation.}
\label{table:sanity}
\begin{tabular}{cccc} \toprule
Window size & Anomalous sequence & \begin{tabular}[c]{@{}c@{}}Flipped prediction \\ after masking\end{tabular} & Percentage flipped \\ \hline
1 & 1511 & 1104 & 73.0\% \\ \hline
3 & 1511 & 1319 & 87.3\% \\ \bottomrule
\end{tabular}
\vspace{-3mm}
\end{table}

\begin{figure*}[b!] 
\centering
\includegraphics[width=0.9\textwidth]{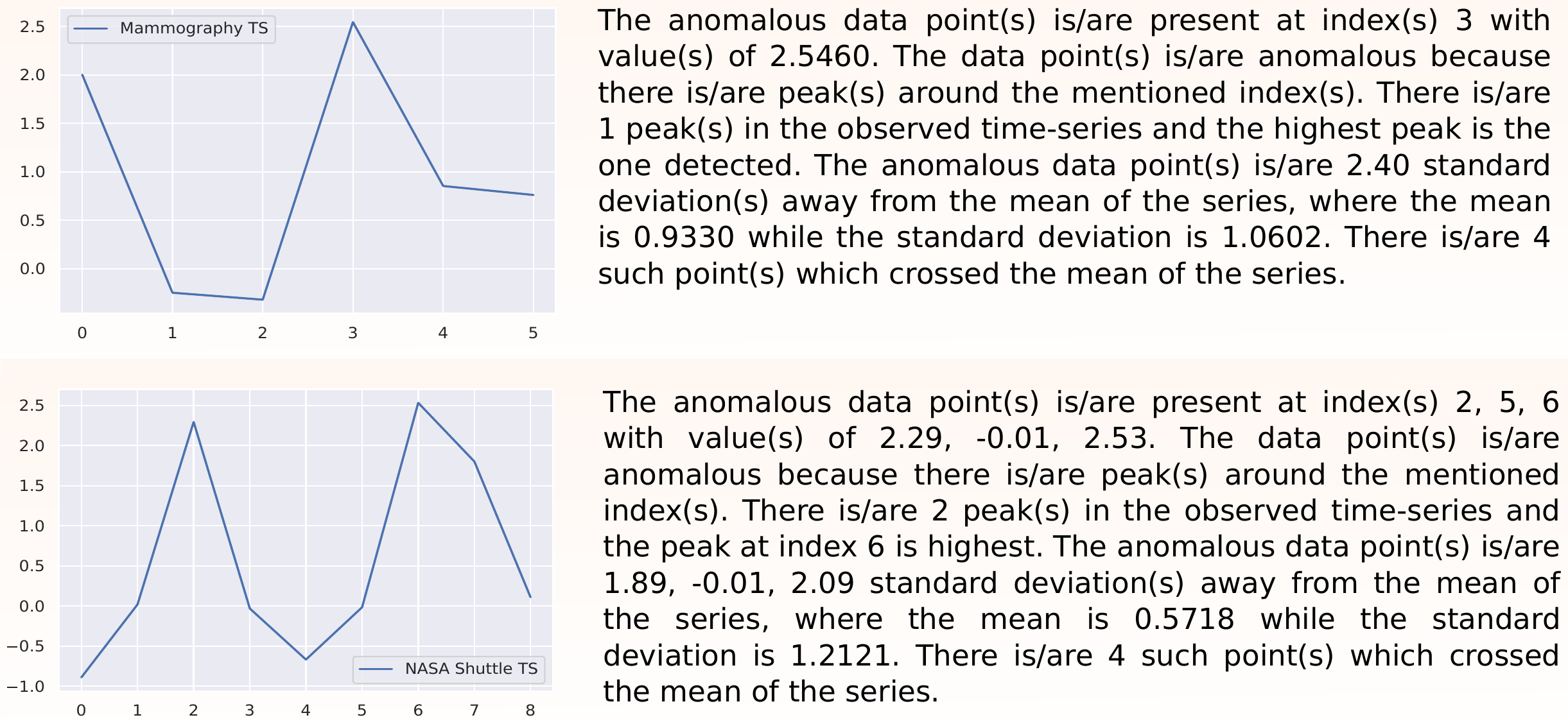}
\caption{Sample explanations generated for Mammography (top) and NASA Shuttle (bottom) time-series.}
\label{fig:mamo_nasa}
\vspace{-5mm}
\end{figure*}

In the traditional interpretation settings where only visual explanations are available, it is difficult for a user to understand why a particular decision is taken by the network just by looking at the plots. However, it is relatively easy to understand the classification reason by reading the explanations provided by our system for the corresponding plots in Fig. \ref{text} and Fig. \ref{fig:mamo_nasa}. We specifically opted for statistical features due to their strong theoretical foundations and transparency. The point-wise features of an anomalous data point help a user in understanding how that data point is different from the rest of the sequence. Whereas sequence-wise features help in highlighting the overall behavior of an anomalous sequence. An end-user can confirm part of the explanation by looking at the plot, which elevates his trust on the system. 

One of the major limitations of the TSXplain is its specificity for the task at hand. The computed features and the rule-base are not easily transferable between different tasks. Another limitation is regarding the availability of a suitable set of statistical features for a particular task. We leave these open questions as future work. We would also like to test our system on more complex datasets along with the employment of more sophisticated influence tracing techniques~\cite{bach2015pixel} to further enhance the utility of the system. 

\vspace{-2mm}
\section{Conclusion} \label{sec:conclusion}
\vspace{-2mm}
This paper proposes a novel demystification framework which generates natural language based descriptions of the decision made by the deep learning models developed for time-series analysis. The \textit{influence tracer} identifies the most salient regions of the input. Statistical features from the sequence are simultaneously extracted from the \textit{statistical feature extractor} module. These two results are passed onto the \textit{textual explanation generator} for the generation of the final explanation by the system. We used statistical features over other alternates due to their strong theoretical foundations and transparency which significantly improved the intelligibility and reliability of the provided explanations. 
A confidence estimate is also provided by the system using the sanity check module which is based on assessment whether the estimated influential point is indeed causal for the prediction. 
The generated explanations are directly intelligible for both expert and novice users alike evading the requirement of domain expertise to understand the encapsulated information. We tested and generated the proposed framework on different synthetic and real anomaly detection datasets and demonstrated our results. The explanations can help the users in uplifting their confidence regarding the performance of the deep model. 
We strongly believe that natural language based descriptions are one of the most convincing ways in the long-term for the realization of explainable systems.

\end{document}